\documentclass[11pt,a4paper]{article}
\usepackage[hyperref]{naaclhlt2019}
\usepackage{hyperref}
\usepackage{url}
\usepackage{amsmath}
\usepackage{amssymb}
\usepackage{algorithm}
\usepackage{graphicx} 
\usepackage{subfigure}
\graphicspath{ {Figure/} }
\usepackage[noend]{algpseudocode}
\usepackage{multirow}
\usepackage{wrapfig}
\usepackage{lipsum}
\usepackage{float}
\graphicspath{ {Figure/} }
\usepackage{times}
\usepackage{latexsym}
\usepackage{url}

\aclfinalcopy 

\title{Reinforcement Learning Based Text Style Transfer without Parallel Training Corpus}

\author{Hongyu Gong\textsuperscript{*}\quad Suma Bhat\textsuperscript{*}\quad Lingfei Wu\textsuperscript{$\dagger$}\quad Jinjun Xiong \textsuperscript{$\dagger$}\quad Wen-mei Hwu\textsuperscript{*} \\
\textsuperscript{*}University of Illinois at Urbana-Champaign, USA \\ \textsuperscript{$\dagger$}T. J. Watson Research Center, IBM \\
\textsuperscript{*}{\{hgong6, spbhat2, w-hwu\}@illinois.edu}\quad  \textsuperscript{$\dagger$}{\{wuli, jinjun\}}@us.ibm.com}

\date{}

\begin{document}
\maketitle
\begin{abstract}

Text style transfer rephrases a text from a source style (e.g., informal) to a target style (e.g., formal) while keeping its original meaning. Despite the success existing works have achieved using a parallel corpus for the two styles, transferring text style has proven significantly more challenging when there is no parallel training corpus. In this paper, we address this challenge by using a reinforcement-learning-based generator-evaluator architecture. Our generator employs an attention-based encoder-decoder to transfer a sentence from the source style to the target style. Our evaluator is an adversarially trained style discriminator with semantic and syntactic constraints that score the generated sentence for style, meaning preservation, and fluency. Experimental results on two different style transfer tasks (sentiment transfer and formality transfer) show that our model outperforms state-of-the-art approaches. Furthermore, we perform a manual evaluation that demonstrates the effectiveness of the proposed method using subjective metrics of generated text quality.

\end{abstract}

\section{Introduction}
Text style transfer is the task of rewriting  a piece of text to a particular style while retaining the  meaning of the original text. It is a challenging task of natural language generation and is at the heart of  many recent NLP applications, such as personalized responses in dialogue system \cite{DBLP:conf/aaai/ZhouLCLCH17}, 
formalized texts \cite{rao2018dear}, 
cyberspace purification by rewriting offensive texts \cite{niu2018polite,santos2018fighting},
and  poetry generation \cite{DBLP:conf/emnlp/YangSYL18}.

Recent works on supervised style transfer with a parallel corpus have demonstrated considerable success \cite{DBLP:journals/corr/JhamtaniGHN17,rao2018dear}.
However, a parallel corpus may not always be available for a  transfer task. This has prompted studies on style transfer without parallel corpora.  These hinge on  the common  idea of separating the content
from the style of the text \cite{shen2017style,DBLP:conf/aaai/FuTPZY18,santos2018fighting}. This line of research first encodes the context via a style-independent  representation, and then transfers sentences by combining the encoded content with style information. In addition, an appropriate training loss is chosen to change the style while preserving the content.
However, these approaches are limited by their use of loss functions that must be differentiable with respect to the model parameters, since they rely on gradient descent to update the parameters. Furthermore, since focusing only on semantic and style metrics in style transfer, they ignore other important aspects of quality in text generation, such as language fluency.

In this paper, we propose a system trained using reinforcement-learning (RL) that performs text style transfer without accessing to a parallel corpus. 
Our model has a generator-evaluator structure with one generator and one evaluator with multiple modules. The generator takes a sentence in a source style as input and transfers it to the target style. It is an attention-based sequence-to-sequence model, which is widely used in generation tasks such as machine translation \cite{luong2015effective}. More advanced model such as graph-to-sequence model can also exploited for this generation task \cite{xu2018graph2seq}. 
The evaluator consists of a style module, a semantic module and a language model for evaluating the transferred sentences in terms of style, semantic content, and fluency, respectively. Feedback from each evaluator is sent to the generator so it can be updated to improve the transfer quality. 

Our style module is a style discriminator built using a recurrent neural network, predicting the likelihood that the given input is in the target style. We train the style module adversarially to be a target style classifier  while regarding the transferred sentences as adversarial samples. An adversarial training renders style classification more robust and accurate. As for the semantic module, we used word movers' distance (WMD), a state-of-the-art unsupervised algorithm for comparing semantic similarity between two sentences \cite{kusner2015word,wu2018word}, to evaluate the semantic similarity between  input sentences in the source style and the transferred sentences in the target style. We also engaged a language model to evaluate the fluency of the transferred sentences. 

Unlike prior studies that separated content from style to guarantee content preservation and transfer strength, we impose explicit semantic, style and fluency constraints on our transfer model. 
Moreover, employing RL allows us to use other evaluation metrics accounting for the quality of the transferred sentences, including non-differentiable ones. 

We summarize our contributions below:\\
\noindent (1) We propose an RL framework for text style transfer. It is versatile to include a diverse set of evaluation metrics as the training objective in our model. \\
\noindent (2)  Our model does not rely on the availability of a parallel training corpus, thus addressing the important challenge of lacking parallel data in many transfer tasks. \\
\noindent (3) The proposed model achieves state-of-the-art performance in terms of content preservation and transfer strength in text style transfer.

The rest of our paper is organized as follows: we discuss related works on style transfer in Section~\ref{sec:related_works}. The proposed text style transfer model and the reinforcement learning framework is introduced in Section~\ref{sec:model}. Our system is empirically evaluated on sentiment and formality transfer tasks in Section~\ref{sec:exp}. We report and discuss the results in Section~\ref{sec:results} and Section~\ref{sec:discussion}. The paper is concluded in Section~\ref{sec:conclusion}.

\section{Related Works}
\label{sec:related_works}


Text generation is one of the core research problems in computational linguistics \cite{song2018graph}, and text style transfer lies in the area of text generation.
Text style transfer has been explored in the context of a variety of natural language applications, 
including sentiment modification \cite{zhang2018learning}, 
text simplification \cite{zhang2017sentence}, 
and personalized dialogue \cite{DBLP:conf/aaai/ZhouLCLCH17}.
Depending on whether the parallel corpus is used for training, two broad classes of style transfer methods have been proposed to transfer the text from the source style to the target style. We will introduce each line of research in the following subsections.

\noindent\textbf{Style transfer with parallel corpus}. Style transfer with the help of a style parallel corpus can be cast as a monolingual machine translation task. For this, a sequence-to-sequence (seq2seq) neural network has been successfully applied in a  supervised setting. \citeauthor{jhamtani2017shakespearizing} transfer modern English to Shakespearean English by enriching a seq2seq model with a copy mechanism to replicate the source segments in target sentences \cite{jhamtani2017shakespearizing}.

\noindent\textbf{Style transfer without parallel corpus}. 
Scarce parallel data in many transfer tasks has prompted a recent interest in studying style transfer without a parallel corpus (e.g., \cite{zhang2018shaped}). \citeauthor{li2018delete} propose to delete words associated with the source style and replace them with similar phrases associated with the target style. Clearly, this approach is limited to transfers at the lexical level and may not handle structural transfer. Most existing unsupervised approaches share a core idea of disentangling content and style of texts. For a given source sentence, a style-independent content representation is firstly derived. Then, in combination with the target style, the content representation is used to generate the sentence following the target style. 

Approaches to extract the content include variational auto-encoders (VAE) and cycle consistency. VAEs are commonly used to learn the hidden representation of inputs for dimensionality reduction,  and have been found to be useful for representing the content of the source  \cite{DBLP:journals/corr/HuYLSX17,mueller2017sequence,shen2017style,DBLP:conf/aaai/FuTPZY18}.
Cycle consistency is an idea borrowed from image style transfer for content preservation \cite{DBLP:conf/iccv/ZhuPIE17}. It proposes to reconstruct the input sentence from the content representation, by forcing the model to keep the information of the source sentence \cite{santos2018fighting}.

The transferred sentences are generated based on the content representation and the target style. One way to achieve this is with the use of  a pre-trained style classifier. The classifier scores the transfer strength of the generated sentences and guides the model to learn the target text style \cite{santos2018fighting,P18-1080}. Another way is to learn the style embedding, which can be concatenated with the content embedding as the representation of the target sentence \cite{DBLP:conf/aaai/FuTPZY18}. The decoder then constructs the sentences from their hidden representations.

We note that previous works rely on gradient descent in their model training, and therefore their training losses (e.g., content and style loss) were limited to functions differentiable with respect to model parameters. Also, very few works consider other aspects of transfer quality beyond the content and the style of the generated sentences. This is in part due to their reliance on a differentiable training objective. We propose an RL-based style transfer system so that we can incorporate more general evaluation metrics in addition to  preserving the semantic meaning of content and style transfer strength.

\noindent\textbf{Reinforcement learning}. RL has recently been applied to challenging NLP tasks \cite{DBLP:conf/aaai/YuZWY17}. RL has advantages over supervised learning in that it supports non-differentiable training objectives and does not need annotated training samples. Benefits of using RL have been demonstrated in image captioning \cite{DBLP:conf/emnlp/GuoCYB18}, sentence simplification \cite{zhang2017sentence}, machine translation \cite{wu2018study} and essay scoring \cite{DBLP:conf/emnlp/WangWZH18}. {A recent work on the task of sentiment transfer  applied reinforcement learning to handle its  BLEU score-based training loss (a non-differentiable function) \cite{xu2018unpaired}. Similar to the style transfer works discussed above, it also disentangled the semantics and the sentiment of sentences using a neutralization module and an emotionalization module respectively.
Our work is different from these related works in that the semantic preservation and transfer strength are taken care of by the use of discriminators without  explicitly separating content and style. An additional aspect that we focus here is the notion of fluency of the transferred sentences, which has not been explored before.}

\section{Model}
\label{sec:model}
\begin{figure}[htbp!]
\centering
\includegraphics[width=0.95\linewidth]{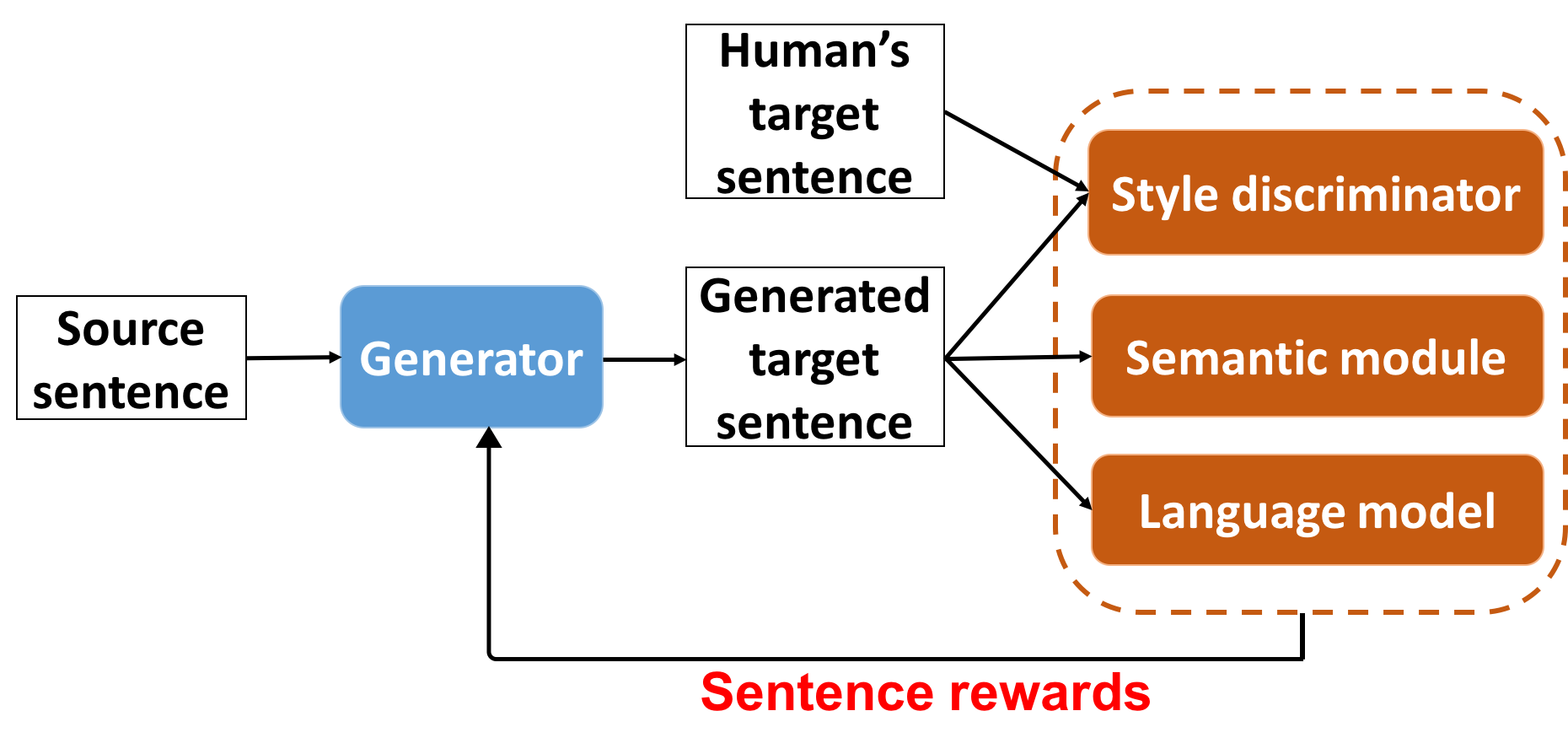}
\caption{Model overview: the generator transfers the input source sentence to the generated target sentence. The generated sentences are collectively evaluated by the style discriminator, the semantic module and language module respectively. The style discriminator is adversarially trained with both human- and model-generated sentences. These three modules evaluate the generated sentences in terms of transfer strength, content preservation and fluency, and the rewards are sent to train the generator.}
\label{fig:overview}
\vspace{-1mm}
\end{figure}

\begin{figure*}[htbp!]
\centering
\includegraphics[width=0.95\linewidth]{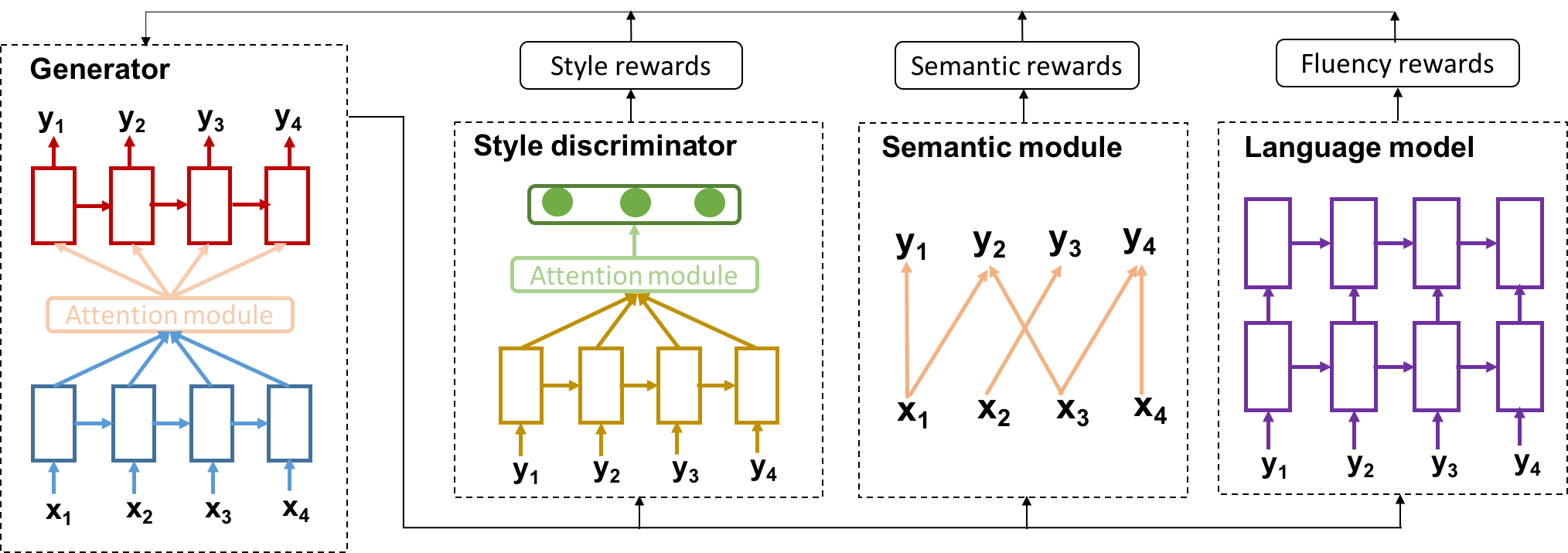}
\vspace{-1mm}
\caption{A detailed view of each component in the text style transfer system.}
\label{fig:detailed}
\vspace{-2mm}
\end{figure*}

Our style transfer system consists of the following  modules: a generator, a style discriminator, a semantic module and a language model as shown in Fig.~\ref{fig:overview}. We next describe the  structure and function of each component. A closer view of our system is presented in Fig.~\ref{fig:detailed}.

\textbf{Generator}. The generator in our system takes a sentence in the source style as input and transfers it to the target style. For this, we use a recurrent encoder-decoder model combined with attention mechanism, which can handle variable-length input and output sequences \cite{sutskever2014sequence,cho2014learning}. We could also leverage recently proposed more advanced encoder-decoder models \cite{xu2018graph2seq,xu2018exploiting} to exploit rich syntactic information for this task, which we leave it as future work. 
Both the encoder and the decoder are recurrent neural layers with gated recurrent units (GRU). The encoder takes one word from the input at each time step, and outputs a hidden state vector $\bar{h}_{s}$ at time $s$. Similarly, the decoder outputs a hidden representation $h_{t}$ at time $t$.

Suppose that the input sequence consists of $T$ words $x=\{x_{1}, \ldots, x_{T}\}$, and the generated target sentence $y$ is also a sequence of words $\{y_{1}, \ldots, y_{T'}\}$. We use $\text{vec}(\cdot)$ to denote the embedding of a word.

The gated recurrent unit dynamically updates its state $h_{t}$ based on its previous state $h_{t-1}$ and current input $i_{t}$. Its computation can be abstracted as $h_{t}=\text{GRU}(h_{t-1}, i_{t})$. For the encoder, the input $i_{t}$ is the embedding of the $t$-th input source word,
\begin{align}
\bar{h}_{t} = \text{GRU}(\bar{h}_{t-1}, \text{vec}(x_{t})).
\end{align}
For the decoder, the input to the recurrent unit is the embedding of the $t$-th generated target word.
\begin{align}
h_{t} = \text{GRU}(h_{t-1}, \text{vec}(y_{t})).
\end{align}

An attention mechanism is commonly adopted in text generation, such as machine translation \cite{bahdanau2014neural,luong2015effective}. We apply the attention mechanism to the decoding step so that the decoder learns to attend to source words and generates words. In this work, we use the attention mechanism similar to that used in \cite{luong2015effective}. At the $t$-th decoding step, the attention $\alpha_{t}(s)$ is the weight of the $s$-th encoder state $\bar{h}(s)$.

The encoder hidden states are linearly weighted by the attention as the context vector at time $t$.
\begin{align}
c_{t} = \sum\limits \alpha_{t}(s)\bar{h}_{s}.
\end{align}
Combining the attention over the source sentence, the decoder produces a new hidden state $\tilde{h}_{t}$,
\begin{align}
\tilde{h}_{t} = \text{tanh}(W_{c}[c_{t};h_{t}]).
\end{align}
The hidden vector $\tilde{h}_{t}$ is then used to predict the likelihood of the next word in the target sentence over the target vocabulary.
\begin{align}
\mathbb{P}(y_{t}|y_{<t},x) = \text{softmax}(W_{s}\tilde{h}_{t}).
\end{align}
where $W_{c}$ and $W_{s}$ are decoder parameters.

\textbf{Style discriminator}. The style discriminator evaluates how well the generated sentences are transferred to the target style. It is a classifier built on a bidirectional recurrent neural network with attention mechanism. The style discriminator is pre-trained to minimize the cross-entropy loss in the style classification task. This style classifier predicts the likelihood that an input sentence is in the target style, and the likelihood is taken as the style score of a sentence.

The pre-training does not guarantee that the neural network model will learn robust style patterns. So we resort to adversarial training as done in  generative adversarial networks (GAN) \cite{DBLP:conf/aaai/YuZWY17,DBLP:conf/emnlp/WangL18a}.
Accordingly, the style discriminator is later adversarially trained to distinguish the original (human-written) sentences from the model-generated ones so that the classifier learns to classify the text style well.

\textbf{Semantic module}. This evaluates how well the  content from the input is preserved in the generated sentences. We use word mover's distance (WMD), which is the state-of-the-art approach (known for its robustness and efficiency) to measure the dissimilarity between the input and output sentences based on word embeddings \cite{kusner2015word,wu2018word}. We take the negative of the WMD distance and divide it by the sequence length to yield the \emph{semantic score} of a generated sentence. Previous works have also used cycle reconstruction loss to measure content preservation by reconstructing input sentences from generated sentences \cite{santos2018fighting}. 

\textbf{Language model}.  The style and the semantic modules do not guarantee the fluency of the transferred sentences. This fluency is achieved using a language model. The language model we use is a two-layer recurrent neural network  pre-trained on the corpus in the target style so as to maximize the likelihood of the target sentences \cite{mikolov2010recurrent,jozefowicz2016exploring}. The language model estimates the probability of input sentences. We take the logarithm of the probability and divide it by the sequence length as the \emph{fluency score}.

\subsection{Reinforcement Learning}
The output sentences from the generator are sent to the  semantic, style and language model modules for evaluation. These  modules give feedback to the generator for the purpose of tuning it and to improve the quality of the generated sentences. We emphasize that despite the fact that our chosen evaluation metrics are not differentiable with respect to the generator parameters, they are still usable here. This is made possible by our use of the RL framework (the REINFORCE algorithm) to update the parameters of the generator \cite{williams1992simple}.

In the RL framework, we define the state and the action for our style transfer task as follows. The state $s_{t}$ at time $t$ is the input source sequence and the first $t-1$ words that are already generated in the target sequence, i.e., $s_{t}=(X, Y_{1:t-1})$. The action $a_{t}$ at time $t$ is the $t$-th word to be generated in the output sequence, i.e., $a_{t}=y_{t}$. Suppose that the target vocabulary is $V$, and the maximum length of the decoder is $T'$.
The generator $G$ is parameterized with a parameter set $\theta$, and we define the expected reward of the current generator as $J(G_{\theta})$. The total expected reward is
\begin{align}
J(G_{\theta})=\sum\limits_{t=1}^{T'}\mathbb{E}_{Y_{1:t-1}\sim G_{\theta}}[\sum\limits_{y_{t}\in V}\mathbb{P}_{\theta}(y_{t}|s_{t})Q(s_{t},y_{t})],
\end{align}
where $\mathbb{P}_{\theta}(y_{t}|s_{t})$ is the likelihood of word $y_{t}$ given the current state, and $Q(s_{t},y_{t})$ is the cumulative rewards that evaluate the quality of the sentences extended from $Y_{1:t}$. Suppose that $r(s_{t},y_{t})$ is the reward of word $y_{t}$ at state $s_{t}$. The total reward, $Q$, is defined as the sum of the word rewards.
\begin{align}
\label{eq:cumulative_reward_def}
Q(s_{t},y_{t})=\sum\limits_{\tau=t}^{T'}\gamma^{\tau-t}r(s_{\tau},y_{\tau}),
\end{align}
where $\gamma$ $(0<\gamma<1)$, is a discounting factor so that the future rewards have decreasing weights,  since their estimates are less accurate.

If we only consider one episode, i.e., $Y_{1:t-1}$ has been given for every $y_{t}$, the reward $J(G_{\theta})$ can be written as 
\begin{align}
\label{eq:total_reward}
J(G_{\theta}) = \sum\limits_{t=1}^{T'}\sum\limits_{y_{t}\in V}\mathbb{P}_{\theta}(y_{t}|s_{t})Q(s_{t},y_{t}).
\end{align}

\textbf{Sequence sampling}. By design, the three evaluation modules in Fig.~\ref{fig:overview} only evaluate complete sentences instead of single words or partial sentences. This means that we cannot obtain $r(s_{t},y_{t})$ directly from the evaluation modules at any time instance before the end of the sentence. One way around this problem  is rolling out \cite{DBLP:conf/aaai/YuZWY17}, where the generator `rolls out' the given sub-sentence $Y_{1:t}$ at time step $t$ to generate complete sentences by sampling the remaining part of the sentence $\{Y_{t+1:T'}^{n}\}$.

Previous works have adopted different sampling strategies, including Monte Carlo search, multinomial sampling and beam search. 
Starting from the given segment $Y_{1:t}$, Monte Carlo search explores the sub-sequence which leads to the best complete sentence \cite{DBLP:conf/aaai/YuZWY17}.  This leads to a good estimate of the sentence rewards but comes at significant computational cost. In many applications, the other two sampling strategies have been adopted for their efficiency.
In multinomial sampling,  each word $y_{\tau}$ $(t<\tau\leq T')$ is sampled from the vocabulary according to the likelihood $\mathbb{P}(y_{\tau}|s_{\tau})$ predicted by the generator   \cite{o2016pgq,chatterjee2010minimum}. The beam search process, on the other hand, keeps track of the $k$ (a user-specified parameter) most likely words at each decoding step rather than just one word \cite{wu2018study}. 
While this yields an accurate estimate of the reward for each action,  multinomial sampling allows us to explore the diversity of generated texts with a potentially higher reward later on. This is the trade-off between exploitation and exploration in RL.

To balance the estimation accuracy and the generation diversity, we combine the ideas of beam search and multinomial sampling. Given a source sentence, we first generate a reference target sentence $Y_{1:T}^{\text{ref}}$ using beam search. To estimate the reward at each time step $t$, we draw samples of complete sentences $\{Y_{1:T'}^{l}\}$ by rolling out the sub-sequence $Y_{1:t}^{\text{ref}}$ using multinomial sampling. The evaluation scores of the sampled sentences are used as reward $r(y_{t},s_{t})$. More details about the sampling process are in Appendix.

\textbf{Reward estimation}. We estimate the reward as follows. We draw $N$ samples of complete sentences starting from $Y_{1:t}$: $\{Y_{1:T'}^{(n)}\}_{n=1}^{N}$. The complete sentences are then fed into the three evaluation modules. Let $f_{\text{style}}$ be the style score given by the  style module, $f_{\text{semantic}}$ be semantic score by the semantic module, and $f_{\text{lm}}$ be the fluency score given by the language model. 

We score the action $y_{t}$ at state $s_{t}$ by the average score of the complete sentences rolled out from $Y_{1:t}$. This action score is defined as the weighted sum of the scores given by the three modules.
\begin{align}
\nonumber
&f(s_{t},y_{t}) = \frac{1}{N}\sum\limits_{n=1}^{N}\left(\right.\alpha\cdot f_{\text{style}}(Y_{1:T'}^{(n)}) + \\
&\beta\cdot f_{\text{semantic}}(Y_{1:T'}^{(n)}, Y_{1:T'}^{\text{real}}) + \eta\cdot f_{\text{lm}}(Y_{1:T'}^{(n)})\left)\right.,
\end{align}
where the hyperparameters $\alpha,\beta$ and $\eta$ are positive. In our experiments, we set $\alpha=1.0$, $\beta=0.5$ and $\eta=0.5$ heuristically.

Given the scores from the evaluation modules, we define the reward $r(s_{\tau},y_{\tau})$ of word $y_{\tau}$ at state $s_{\tau}$ as
\begin{align}
\label{eq:single_reward}
r(s_{\tau},y_{\tau})=\left\{
\begin{aligned}
&f(s_{\tau},y_{\tau})-f(s_{\tau-1},y_{\tau-1}),&\tau>1,\\
&f(s_{1},y_{1}),&\tau=1.
\end{aligned}
\right.
\end{align}

We then obtain the discounted cumulative reward $Q(s_{t},y_{t})$ from the rewards $\{r(s_{\tau},y_{\tau})\}_{\tau>t}$ at each time step using Eq.~\ref{eq:cumulative_reward_def}.

The total reward of $J(G_{\theta})$ can be derived from the cumulative rewards $\{Q(s_{t},y_{t})\}$ using Eq.~\ref{eq:total_reward}. We define the generator loss $L_{\theta}$ as the negative of reward $J(G_{\theta})$, $L_{G}(\theta)=-J(G_{\theta})$.

According to Eq.~\ref{eq:total_reward}, we can find the gradient $\nabla_{\theta}L_{\theta}$ of the generator loss as,
\begin{align}
\nabla_{\theta}L_{G}(\theta)=-\sum\limits_{t=1}^{T'}\nabla_{\theta}\mathbb{P}_{\theta}(y_{t}|s_{t})Q(s_{t},y_{t}).
\end{align}

\subsection{Adversarial Training}
The style discriminator is pre-trained on corpora in the source and target styles, and is used to evaluate the strength of style transfer. We note that this pre-training may not be sufficient for the style classifier to learn robust patterns and to provide accurate style evaluation. 
Indeed, in our experiments we found that even though the generator was trained to generate target sentences by maximizing the style rewards, the  one-shot pre-training was insufficient to render the sentences  in the target style. 

Borrowing the idea of adversarial training proposed in GANs, we continuously trained the style discriminator using the generated target sentences. Toward this, we used a combination of a randomly sampled set of human-written target sentences $\{Y_{\text{human}}^{(k)}\}$ and   model-generated sentences $\{Y_{\text{model}}^{(k)}\}$. Here the model-generated instances act as adversarial training samples, using which, the style discriminator was trained to distinguish the model outputs from human-written sentences. Let the discriminator $D$ be parameterized by a parameter set $\phi$. We define  the prediction of the style discriminator, $D(Y)$, as the likelihood that the sentence $Y$ is in the target style. The objective of this adversarial training amounts to  minimizing the discriminator loss $L_{D}$:
\begin{align}
\nonumber
L_{D}(\phi) &= \frac{1}{K}\left(\right.-\sum\limits_{k=1}^{K}\log(1-D_{\phi}(Y_{\text{model}}^{(k)}))\\
&-\sum\limits_{k=1}^{K}\log D_{\phi}(Y_{\text{human}}^{(k)})\left)\right..
\end{align}

\section{Experiments}
\label{sec:exp}
In this work, we considered two textual style transfer tasks, that of sentiment transfer (\textbf{ST}, involving negative and positive sentiments) and formality transfer (\textbf{FT}, involving informal and formal styles) using two curated datasets.  We experimented with both transfer directions: positive-to-negative, negative-and-positive, informal-to-formal and formal-to-informal.

\noindent\textbf{Dataset}. 
For our experiments with style transfer we used a sentiment corpus and a formality corpus described below. \\

\begin{table}[htbp!]
\vspace{-2mm}
\centering
\resizebox{0.48\textwidth}{!}{
\begin{tabular}{|c|c|c|c|c|c|}
\hline
& Vocabulary & Type & Train & Dev & Test \\ \hline
\multirow{2}{*}{Sentiment} & \multirow{2}{*}{9,640} & Negative & 176,878 & 25,278 & 50,278 \\ \cline{3-6}
 & & Positive & 267,314 & 38,205 & 76,392 \\ \hline
\multirow{2}{*}{Formality}& \multirow{2}{*}{21,129} & Informal & 50,711 & 1,019 & 1,327 \\ \cline{3-6} 
 & & Formal & 50,711 & 1,019 & 1,019 \\ \hline
\end{tabular}}
\vspace{-1mm}
\caption{Data sizes of sentiment and formality transfer.}
\label{tab:data_size}
\vspace{-0mm}
\end{table}

\noindent(1) {Sentiment corpus}. The sentiment corpus consists of restaurant reviews collected from the Yelp website \cite{shen2017style}. The reviews are classified as either negative or positive. \\
\noindent(2) {Formality corpus}. We use the  Grammarly's Yahoo Answers Formality Corpus (GYAFC) \cite{rao2018dear}, which is a collection of sentences posted in  a question-answer forum (Yahoo Answers) and written in an informal style. In addition, these sentences have been manually rewritten in a formal style. We used the data from the section  \textit{family and relationships}. Note that even though the corpus is parallel, we did not use the parallel information.

Table~\ref{tab:data_size} shows the train, dev and test data sizes as well as the vocabulary sizes of the corpora used in this work.

\begin{table*}[htbp!]
\resizebox{\textwidth}{!}{
\begin{tabular}{|l|l|l|}
\hline
Type & Source sentence & Transferred sentence \\ \hline
Negative-to-Positive & \begin{tabular}[c]{@{}l@{}}Crap fries , hard hamburger buns ,\\burger tasted like crap !\end{tabular} & \begin{tabular}[c]{@{}l@{}}Love you fries, burgers , always fun burger , \\authentic !\end{tabular} \\ \hline
Positive-to-Negative & I was very impressed with this location . & I was very disappointed with this location . \\ \hline
Informal-to-Formal & \begin{tabular}[c]{@{}l@{}}It defenitely looks like he has feelings \\ for u do u show how u feel u should ! !\end{tabular} & \begin{tabular}[c]{@{}l@{}}It is like he is interested in you you should \\ show how you feel .\end{tabular} \\ \hline
Formal-to-Informal & \begin{tabular}[c]{@{}l@{}}I believe you 're a good man most \\ likely she loves you quite a bit.\end{tabular} & I think you 're a good man she kinda loves you . \\ \hline
\end{tabular}}
\vspace{-1mm}
\caption{Example transferred sentences.}
\label{tab:example}
\vspace{-2mm}
\end{table*}

\noindent\textbf{Model settings}. The word embeddings used in this work were of dimension 50. They were first trained on the English WikiCorpus and then tuned on the training dataset. The width of the beam search (parameter $k$) was $8$ during the RL and the inference stage. 

\noindent\textbf{Pre-training}. We pre-trained the generator, the style discriminator and the language model before the reinforcement learning stage. We discuss each of these steps below. \\
\noindent\textbf{Generator pre-training}. We pre-trained the generator to capture the target style from the respective target corpus. This pre-training occurred  before setting up the reward from the evaluator to update its parameters in reinforcement learning. During pre-training, we used a set of  target instances with a given  instance serving  as the input as well as  the expected output. Using this set we trained the generator  in a supervised manner with the cross-entropy loss as the training objective. Pre-training offered two immediate benefits for the generator: 
(1) the encoder and decoder learned to capture the semantics and the target style from the target corpus;
(2) the generator had a good set of initial  parameters that led to faster model training. This second aspect is a significant gain, considering that reinforcement learning is more time consuming than supervised learning.\\
\noindent\textbf{Style discriminator pre-training}. The style discriminator in our work was built using a bidirectional recurrent neural network. It was pre-trained using training corpora consisting of sentences in both the source and the target styles. We trained it to classify the style of the input sentences with the cross-entropy classification loss. \\
\noindent\textbf{Language model pre-training}. The language model was a two-layer recurrent neural network. Taking a target sentence $y=\{y_{1}, \ldots, y_{T'}\}$ as the input, the language model predicted the probability of the $t$-th word $y_{t}$ given the previous sub-sequence $y_{1:t-1}$. The language model was pre-trained on the training corpus in target style to maximize the probability of $y_{t}$  $(1\leq t\leq T')$. 

\noindent\textbf{Baselines}. 
We considered two state-of-the-art methods of unsupervised text style transfer that use non-parallel training corpus. \\
\noindent(1) Cross alignment model (CA). The CA model assumes that the text in the source and target style share the same latent content space \cite{shen2017style}. The style-independent content representation generated by its encoder is combined with available style information to transfer the sentences to the target style. 
We used their publicly available  model for ST, and  trained the model  for FT  separately with its default parameters. \\
\noindent(2) Multi-decoder seq2seq model (MDS). MDS consists of one encoder and multiple decoders \cite{DBLP:conf/aaai/FuTPZY18}. Similar to the cross alignment transfer, its encoder learns style-independent representations of the source, and the style specific decoder will rewrite sentences in the target style based on the content representation. We trained the model  with its default parameters for both the tasks.

\subsection{Evaluation}
We used both automatic and human evaluation to validate our system in terms of content preservation, transfer strength and fluency. 

\begin{table*}[htbp!]
\centering
\resizebox{0.8\textwidth}{!}{
\begin{tabular}{|c|c|c|c|c|c|c|c|c|}
\hline
Sentiment & \multicolumn{4}{c|}{Negative-to-Positive} & \multicolumn{4}{c|}{Positive-to-Negative} \\ \hline
Metric & Content & Style & Overall & Perplexity & Content & Style & Overall & Perplexity \\ \hline
CA & \textbf{0.894} & 0.836 & 0.432 & 103.11 & 0.905 & 0.836 & 0.435 & 185.35 \\ \hline
MDS & 0.783 & 0.988 & 0.437 & \textbf{98.89} & 0.756 & 0.860 & 0.402 & \textbf{156.98} \\ \hline
RLS & 0.868 & 0.98 & \textbf{0.460} & 119.24 & 0.856 & 0.992 & \textbf{0.459} & 174.02 \\ \hline
Formality & \multicolumn{4}{c|}{Informal-to-Formal} & \multicolumn{4}{c|}{Formal-to-Informal} \\ \hline
Metric & Content & Style & Overall & Perplexity & Content & Style & Overall & Perplexity \\ \hline
CA & 0.865 & 0.558 & 0.339 & 238.05 & 0.789 & 0.956 & 0.432 & 317.40 \\ \hline
MDS & 0.519 & 0.435 & 0.237 & 278.65 & 0.546 & 0.998 & 0.353 & 352.86 \\ \hline
RLS & 0.885 & 0.601 & \textbf{0.358} & \textbf{208.33} & 0.873 & 0.982 & \textbf{0.462} & \textbf{267.78} \\ \hline
\end{tabular}}
\vspace{-1mm}
\caption{Automatic evaluation of text style transfer systems on sentiment and formality transfer.}
\label{tab:auto_result}
\vspace{-2mm}
\end{table*}

\subsubsection{Automatic evaluation}
Aligning with prior work, we used the automatic metrics of content preservation, transfer and fluency that have been found to be well correlated with human judgments  \cite{DBLP:conf/aaai/FuTPZY18}. For comparison, in Appendix, we also report  our style and semantic metrics as provided by the evaluator.

\noindent\textbf{Content preservation}.  A key requirement of the transfer process is that the original meaning be retained. Here we measure this by  an embedding based sentence similarity metric $s_{\text{sem}}$ proposed by \cite{DBLP:conf/aaai/FuTPZY18}. 
The embedding we used was based on the word2vec (CBOW) model \cite{DBLP:conf/nips/MikolovSCCD13}. It was first trained on the English WikiCorpus and then tuned on the training dataset. Previous works used pre-trained GloVe embedding \cite{DBLP:conf/emnlp/PenningtonSM14}, but we note that it does not have embeddings for Internet slang commonly seen in sentiment and formality datasets. 

\noindent\textbf{Transfer strength}. The transfer strength $s_{\text{style}}$ captures the degree to which the style transfer was carried out and was  quantified using a classifier. An LSTM-based classifier was trained for style classification on a training corpus \cite{DBLP:conf/aaai/FuTPZY18}. The classifier predicts the style of the generated sentences with a threshold of 0.5. The prediction accuracy is defined as the percentage of generated sentences that were classified to be in the target style. The accuracy was used to evaluate transfer strength, and the higher the accuracy is, the better the generated sentences fit in target style.

\noindent\textbf{Overall score}. We would like to point out that there is a trade-off between content preservation and transfer strength. This is because the outputs resulting from unchanged input sentences show the best content preservation while  having poor  transfer strength. Likewise, for given inputs,  sentences sampled from the target 
corpora have the strongest transfer strength while barely preserving any content if at all.
To combine the evaluation of semantics and style, we use the overall score $s_{\text{overall}}$, which is defined as a function of  $s_{\text{sem}}$ and $s_{\text{style}}$:
$s_{\text{overall}} = \frac{s_{\text{sem}} * s_{\text{style}}}{s_{\text{sem}} + s_{\text{style}}}$ \cite{DBLP:conf/aaai/FuTPZY18}.

\begin{table*}[htbp!]
\centering
\resizebox{0.9\textwidth}{!}{
\begin{tabular}{|l|c|c|c|c|}
\hline
\multicolumn{1}{|l|}{Metric} & \multicolumn{1}{l|}{Negative-to-positive} & \multicolumn{1}{l|}{Positive-to-negative} & \multicolumn{1}{l|}{Informal-to-formal} & \multicolumn{1}{l|}{Formal-to-informal} \\ \hline
Content (1-6) & 5.19 & 5.20 & 4.96 & 5.33 \\ \hline
Style accuracy & 0.90 & 0.91 & 0.83 & 0.86 \\ \hline
Fluency (1-6) & 5.51 & 5.61 & 5.33 & 5.21 \\ \hline
\end{tabular}}
\vspace{-1mm}
\caption{Human judgments of transferred sentences}
\label{tab:human_results}
\vspace{-2mm}
\end{table*}

\noindent\textbf{Fluency}. This is usually evaluated with a language model in many NLP applications \cite{DBLP:journals/pdln/PerisC15,DBLP:conf/interspeech/TuskeSN18}. We used a two-layer recurrent neural network with gated recurrent units as a language model, and trained it on the target style part of the corpus. The language model gives an estimation of perplexity (PPL)  over each generated sentence. Given a word sequence of $M$ words $\{w_{1}, \ldots, w_{M}\}$ and the sequence probability $p(w_{1}, \ldots, w_{M})$ estimated by the language model, the perplexity is defined as:
\begin{align}
\text{PPL} = p(w_{1}, \ldots, w_{M})^{-\frac{1}{M}}.
\end{align}
The lower the perplexity on a sentence, the more fluent the sentence is.

\subsubsection{Human annotation}
Noting the best overall score of our system in both directions of the tasks considered (to be discussed in the  section that follows), we performed  human annotations for content, style and fluency to validate the automatic scores. We chose a sample of $100$ sentences generated  by our system for each transfer task and collected three human judgments per sentence in each evaluation aspect. The annotation guidelines were:

\noindent\textbf{Content preservation}. Following the annotation scheme adopted by \cite{rao2018dear}, we asked annotators to rate the semantic similarity between the original and transferred sentence on a scale from 1 to 6. Here ``1'' means completely dissimilar, ``2'' means dissimilar but on the same topic, ``3'' means dissimilar while sharing some content, ``4'' means roughly similar, ``5'' means almost similar, and ``6'' means completely similar.

\noindent\textbf{Transfer strength}. 
Annotators were given pairs of original and transferred sentences and were asked to decide which one was more likely to be in the target style. We define  transfer strength to be the percentage of transferred sentences that were classified to be in the target style.

\noindent\textbf{Fluency}. Similar to the annotation of content, annotators scored sentences for fluency on a scale of 1(\textit{not fluent}) to 6 (\textit{perfectly fluent}). 

\section{Results}
\label{sec:results}
Some example sentences transferred by our system are shown in Table~\ref{tab:example}. More transferred sentences generated by our system and those by the baseline methods can be found in the  Appendix. We first report the results of the automatic evaluation of our proposed system (denoted as ``RLS'') and the two baselines--the cross alignment model (CA) \cite{shen2017style} and the multi-decoder seq2seq model (MDS) \cite{DBLP:conf/aaai/FuTPZY18}--in Table~\ref{tab:auto_result}. 

\noindent\textbf{Sentiment transfer}.  We notice that CA was the best in preserving content, MDS  generated the most fluent target sentences and our model achieved the best trade-off between meaning and style with the  highest overall score. Looking at the Overall score, it is notable that despite the differences in performance between the models studied here, each one performs  similarly in both directions. This could be interpreted to mean that with respect to difficulty of transfer, style transfer is equivalent in both the directions for this task. 

\noindent\textbf{Formality transfer}. For this task, we notice that our model outperforms the baselines in terms of content preservation, transfer strength and fluency with the best Overall score and perplexity. This suggests that our model is better at capturing formality characteristics compared to the baselines.
We also note  that the style strength of all models for informal-to-formal transfer is significantly lower than that for formal-to-informal transfer. This suggests that the informal-to-formal transfer is harder than the reverse. A plausible explanation is that  informal sentences are more diverse and thus easier to generate than formal sentences. For example, informality can be achieved by multiple ways, such as by using an abbreviation (e.g., ``u'' used as ``you'') and adding speech markers (e.g., ``hey'' and ``ummm''), while formality is achieved in a more restricted manner.

Another challenge for informal-to-formal transfer is that informal data collected from online users usually contain non-negligible spelling errors such as ``defenetely'', ``htink'' and ``realy''. Words being the smallest semantic units in all the models considered here, these spelling errors could affect the transfer performance.

For each direction of transfer, we average the scores by annotators for each evaluation item, and report the results in Table~\ref{tab:human_results}. Our transferred sentences are shown to have good quality in content, style and fluency in subjective evaluations. 

\section{Discussion}
\label{sec:discussion}
To gain insights into the ways in which our approach performs the intended style transfer, 
we randomly sampled the generated sentences in the informal-to-formal transfer task. We found that the forms of rewriting can be broadly classified as: lexical substitution, word removal, word insertion and structural change. We show the following examples to these forms of re-writing, where the changed parts are highlighted.\\
\noindent (1) Lexical substitution. The informal sentence ``I do \textbf{n't} know what \textbf{u} mean'' was transferred to ``I do \textbf{not} know what \textbf{you} mean'';\\
\noindent (2) Word removal. The informal sentence ``\textbf{And} I dont know what I should do'' was rewritten as ``I do not know what I should do'';\\
\noindent (3) Word insertion.  In the example instance   ``depends on the woman''  that was changed to ``\textbf{It} depends on the woman'', we see  that a subject was added to generate a complete formal sentence.\\
\noindent (4) Structural change. A small number of instances were also rewritten by making structural changes. For example, the informal sentence ``\textbf{Just} tell them , \textbf{what are they gonna do} , slap you ??'' was transferred to a formal version as ``\textbf{You should} tell them , \textbf{they can not} slap you''.
Other ways  of  style transfer  by incorporating evaluation metrics of structural diversity are left for future work.

\section{Conclusion}
\label{sec:conclusion}
We proposed a reinforcement-learning-based text style transfer system that can incorporate any evaluation metric to enforce semantic, stylistic and fluency constraints on transferred sentences.  We demonstrated its efficacy via automatic and human evaluations using curated datasets  on two different style transfer tasks. We will explore and incorporate other metrics to improve other aspects of generated texts such as the structural diversity in the future work.

\bibliography{naaclhlt2019}
\bibliographystyle{acl_natbib}
\newpage

\appendix
\onecolumn
\section{Appendices}
\label{sec:appendix}

\subsection{Sequence Sampling in Reinforcement Learning}
\label{subsec:app_sample}
\begin{figure}[htbp!]
\centering
\includegraphics[width=0.95\linewidth]{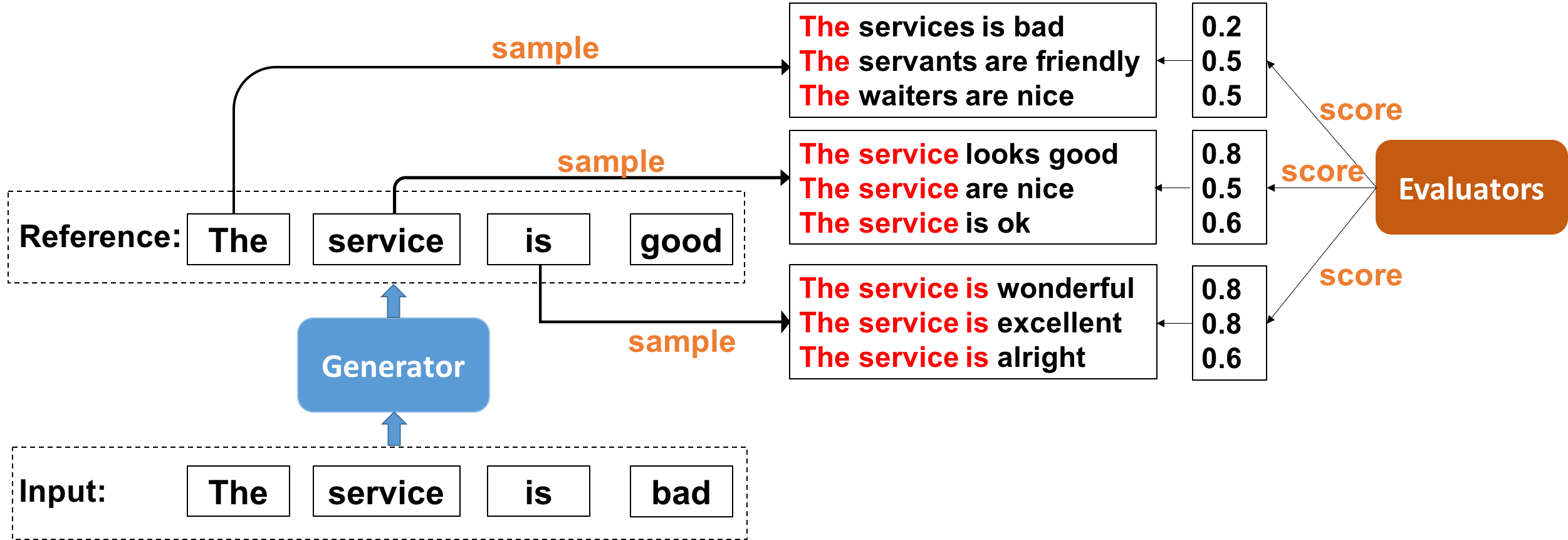}
\caption{Sequence sampling: red words are sub-sequences in reference target sentence based on which the remaining sub-sequences are sampled. The sampled complete sentences are sent to the evaluator for scoring.}
\label{fig:app_sample}
\end{figure}

The generator $G$ transfers a source sentence $X$ into a sentence in target style. In this work, we use beam search of width $k$ to find a reference target sentence $Y_{1:T'}^{\text{ref}}$. In RL, we need to estimate the reward of each action $y_{t}$ in the reference sentence $Y_{1:T'}^{\text{ref}}$. Fig.~\ref{fig:app_sample} shows the sampling and scoring process. 

Suppose that the reference target sentence $Y_{1:T'}^{\text{ref}}$ is ``The service is good''. At the first time step, i.e., when $t=1$, we start from the sub-sequence $Y_{1:1}^{\text{ref}}$, i.e., the sub-sentence "The". We use multinomial sampling to roll out ``The'' to complete sentences, which are ``The service is bad'', ``The servants are friendly'' and ``The waiters are nice'' in Fig.~\ref{fig:app_sample}. These sampled sentences are then sent to the evaluator for scoring in terms of style, content and fluency. Their scores are 0.2, 0.5 and 0.5 respectively, and we average them as the action score $f(y_{1},s_{1})$ of the first word $y_{1}$ at its state $s_{1}$. The score $f(y_{1},s_{1})=0.4$ is sent back to the generator, which will be used to obtain the reward $r(y_{1}, s_{1})$ as described in Eq.~\ref{eq:single_reward}. Similarly when $t=2$, we sample three complete sentences based on the sub-sentence ``The service'':  ``The service looks good'', ``The service are nice'' and  ``The service is ok''.

\subsection{Experiments}
\label{subsec:app_exp}
\begin{table}[htbp!]
\centering
\caption{Semantic and style scores given by our evaluator on all systems.}
\label{tab:app_auto_eval}
\begin{tabular}{ccccccccc}
\hline
\multicolumn{1}{|c|}{} & \multicolumn{4}{c|}{Sentiment} & \multicolumn{4}{c|}{Formality} \\ \hline
\multicolumn{1}{|c|}{} & \multicolumn{2}{c|}{Negative-to-Positive} & \multicolumn{2}{c|}{Positive-to-Negative} & \multicolumn{2}{c|}{Informal-to-Formal} & \multicolumn{2}{c|}{Formal-to-Informal} \\ \hline
\multicolumn{1}{|c|}{Metric} & \multicolumn{1}{c|}{Semantic} & \multicolumn{1}{c|}{Style} & \multicolumn{1}{c|}{Semantic} & \multicolumn{1}{c|}{Style} & \multicolumn{1}{c|}{Semantic} & \multicolumn{1}{c|}{Style} & \multicolumn{1}{c|}{Semantic} & \multicolumn{1}{c|}{Style} \\ \hline
\multicolumn{1}{|c|}{CA} & \multicolumn{1}{c|}{\textbf{-1.293}} & \multicolumn{1}{c|}{0.806} & \multicolumn{1}{c|}{\textbf{-1.346}} & \multicolumn{1}{c|}{0.818} & \multicolumn{1}{c|}{-1.212} & \multicolumn{1}{c|}{0.646} & \multicolumn{1}{c|}{-1.281} & \multicolumn{1}{c|}{0.851} \\ \hline
\multicolumn{1}{|c|}{MDS} & \multicolumn{1}{c|}{-1.412} & \multicolumn{1}{c|}{\textbf{0.855}} & \multicolumn{1}{c|}{-1.662} & \multicolumn{1}{c|}{0.822} & \multicolumn{1}{c|}{-1.508} & \multicolumn{1}{c|}{0.568} & \multicolumn{1}{c|}{-1.445} & \multicolumn{1}{c|}{\textbf{0.878}} \\ \hline
\multicolumn{1}{|c|}{RLS} & \multicolumn{1}{c|}{-1.315} & \multicolumn{1}{c|}{0.846} & \multicolumn{1}{c|}{-1.458} & \multicolumn{1}{c|}{\textbf{0.847}} & \multicolumn{1}{c|}{\textbf{-0.935}} & \multicolumn{1}{c|}{\textbf{0.782}} & \multicolumn{1}{c|}{\textbf{-0.903}} & \multicolumn{1}{c|}{0.872} \\ \hline
\end{tabular}
\end{table}

\noindent\textbf{Automatic evaluation metrics}. We reported the automatic evaluation results of all text style transfer systems in Table~\ref{tab:auto_result}, where we used the evaluation metrics adopted by previous works \cite{DBLP:conf/aaai/FuTPZY18,santos2018fighting}. Here we report the style and semantic scores given by the evaluator in our system in Table~\ref{tab:app_auto_eval}. Recall that semantic score given by our evaluator was the negative of word movers' distance between the generated sentence and the source sentence divided by the sentence length. The larger the semantic score was, the better the content was preserved in the generated sentence. As for the style evaluation, we used a bidirectional recurrent neural network as style classifier. It predicted the likelihood that an input sentence was in target style, which was taken as the \emph{style score} of the generated sentences. Again, the larger the style score was, the better the generated sentence fitted in target style.

As shown in Table~\ref{tab:app_auto_eval}, the results given by the semantic and style modules of our evaluator are very similar to those given by \citeauthor{DBLP:conf/aaai/FuTPZY18}. In sentiment transfer task, CA model does best in content preservation and MDS does best in transfer strength. As for FT, our model outperforms the two baselines in terms of semantic and style scores.

\begin{table}[htbp!]
\caption{Example transferred sentences of all systems.}
\label{tab:app_example}
\resizebox{\textwidth}{!}{
\begin{tabular}{|c|l|}
\hline
Type & \multicolumn{1}{c|}{Transferred sentence} \\ \hline
\multirow{8}{*}{Negative-to-positive} & Source: I 've noticed the food service sliding down hill quickly this year . \\
 & CA: I have enjoyed the food here and this place is perfect . \\
 & MDS: Food is the best staff . \\
 & RLS: I 've noticed the food service was perfect this time . \\ \cline{2-2}
 & Source: The chicken tenders did n't taste like chicken , wtf ? \\
 & CA: The food tastes good , just like spicy ! \\
 & MDS: And the food is the food in the food in well .\\
 & RLS: . The chicken were like chicken, you can find what you want .\\
 \hline
\multirow{4}{*}{Positive-to-negative} & Source: I recommend ordering the `` special chicken '' really good ! \\
 & CA: I would give the pizza ... how they really really good ? \\
 & MDS: They are the worst customer service . \\
 & RLS: I would say chicken were very bad . \\ \cline{2-2}
 & Source: My experience was brief , but very good . \\
 & CA: My experience was ok , but , very good . \\
 & MDS: Worst , i would never go to going back . \\
 & RLS: My experience was bad . \\
 \hline
\multirow{4}{*}{Informal-to-formal} & Source: Well that is just the way it is i guess . \\
 & CA: It is the best thing i think that is not . \\
 & MDS: That is for the way . \\
 & RLS: It is the way I think . \\ \cline{2-2}
 & Source: Like i said he already knows that you like him , so just take a deep breathe and ask him . \\
 & CA: I think that she likes you , but perhaps you will get a relationship and and ask her .\\
 & MDS: If you find him and i think that you have been in a relationship .\\
 & RLS: I believe he knows that you like him, so go to ask him . \\
 \hline
\multirow{4}{*}{Formal-to-informal} & Source: Well, if you are really attracted to this guy, then smile and speak nicely to him . \\
 & CA: If you to tell her the way that is you and get married . \\
 & MDS: The way of guys are not if you are not . \\
 & RLS: Well , if you really like this guy , then smile to him . \\ \cline{2-2}
 & Source: Men are unintelligent! What person understands the meaning behind their behavior? \\
 & CA: Men are not of his meaning .\\
 & MDS: Men are understands all men are not ? \\
 & RLS: Men are stupid ! Why girl loves the mind ? \\
 \hline
\end{tabular}}
\vspace{-0.2cm}
\end{table}

\noindent\textbf{Examples and Analysis}. We list some example transferred sentences given by our model and two baseline systems in Table~\ref{tab:app_example}. In the first example of negative-to-positive transfer, our model adheres to the topic of food service while baselines change to topic of food. Similarly in the first example of positive-to-negative transfer, our model preserves the topic of chicken while CA model talks about pizza and MDS model talks about customer service. Semantic similarity as explicit semantic constraints in our model is shown to be better at preserving the topic of source sentences.

There is still space to improve content preservation in all models. In the second example of informal-to-formal transfer, all transferred sentences miss the segment of ``take a deep breathe'' in the source sentence.
In the second example of formal-to-informal transfer, the three transferred sentences miss part of source information. The source sentence is a rhetorical question, which truly means ``people hardly understand the meaning behind their behavior''. This is a hard example, and all models do not capture its semantic meaning accurately.

FT task is more challenging compared with ST given that the sentence structure is more complicated with a larger vocabulary in the formality dataset. Its difficulty is also reflected by the degraded transfer performance of all systems as reported in Table.~\ref{tab:auto_result}. From the examples in Table~\ref{tab:app_example}, our transferred sentences are more fluent than the outputs of two baselines in FT. The language model in our system plays an important role in making the model's outputs more fluent.


\end{document}